\documentclass{article}

\usepackage{arxiv}

\usepackage[utf8]{inputenc} 
\usepackage[T1]{fontenc}    
\usepackage{hyperref}       
\usepackage{url}            
\usepackage{booktabs}       
\usepackage{amsfonts}       
\usepackage{nicefrac}       
\usepackage{microtype}      
\usepackage{lipsum}

\usepackage{xspace}
\usepackage{color}
\usepackage{arydshln}
\usepackage{booktabs}
\usepackage{pgfplotstable}
\usepackage{tikz}
\usepackage{caption}
\usepackage[utf8]{inputenc}
\usepackage[T1]{fontenc} 
\usepackage{blindtext}
\usepackage{epstopdf}
\usepackage{multirow}
\pgfplotsset{compat=newest}
\usepackage{pdfpages}
\usepackage{float}
\usepgfplotslibrary{units}
\usepackage{subcaption}
\usepackage{lipsum}
\usepackage{algorithmic}
\usepackage{amsmath}

\title{Polylingual Wordnet}

\author{
  Mihael Arcan \\
  Insight Centre for Data Analytics\\
  Data Science Institute\\
  National University of Ireland Galway \\
  \texttt{mihael.arcan@insight-centre.org} \\
   \And
 John P. M\textsuperscript{c}Crae \\
  Insight Centre for Data Analytics\\
  Data Science Institute\\
  National University of Ireland Galway \\
  \texttt{john.mccrae@insight-centre.org} \\
  \And
 Paul Buitelaar \\
  Insight Centre for Data Analytics\\
  Data Science Institute\\
  National University of Ireland Galway \\
  \texttt{paul.buitelaar@insight-centre.org} \\
}

\begin{document}
\maketitle


\section{Introduction}

Princeton WordNet \cite{Fellbaum:1998} is one of the most important resources used in many different tasks across linguistics and natural language processing, however the resource is only available for English and is limited in its coverage of real world concepts. To cross the language barrier, huge efforts have been made to extend the Princeton WordNet with multilingual information in projects, such as EuroWordNet~\cite{vossen1998eurowordnet}, BalkaNet~\cite{tufis2004balkanet} and MultiWordNet~\cite{MultiWordNet_2002}, mostly following the extend approach, where the structure of the Princeton WordNet is preserved and only the words in each synset are translated and new synsets are added for concepts. Furthermore, the Princeton WordNet has many fewer concepts than large scale encyclopedias such as Wikipedia\footnote{\url{https://en.wikipedia.org/wiki/Main_Page}} and resources derived from it such as DBpedia~\cite{auer2007dbpedia} and BabelNet~\cite{Navigli:2012:BAC:2397213.2397579}. This problem is even worse for many non-English wordnets, due to the extend approach, as these resources have even fewer synsets than Princeton WordNet. Furthermore, there are still many languages for which a wordnet does not exist or is not available to all potential users due to licensing restrictions. 

To address these deficiencies we propose two approaches. Firstly, we apply high-quality statistical machine translation (SMT) to automatically translate the WordNet entries into several different European languages. While an SMT system can only return the most frequent translation when given a term by itself, we propose a novel method to provide strong word sense disambiguation when translating wordnet entries. In addition, our method can handle fundamental complexities such as the need to translate all senses of a word including low-frequency senses, which is very challenging for current SMT approaches. For these reasons, we leverage existing translations of Princeton WordNet entries in other languages to identify contextual information for wordnet senses from a large set of generic parallel corpora.  The goal is to identify sentences that share the same semantic information in respect to the synset of the Princeton WordNet entry that we want to translate. Secondly, we describe a novel system based on state-of-the-art semantic textual similarity and ontology alignment to establish a new linking between Princeton WordNet and DBpedia. This method uses a multi-feature approach to establish similarities between synsets and DBpedia entities based on analysis of the definitions using a variety of methods from simple string statistics, to methods based on explicit semantic analysis as well as deep learning methods including long short-term memory (LSTM) networks. These statistics are created based on the Princeton WordNet synset gloss as well as the neighbouring words in the WordNet graph. These are combined using a constraint-based solver that considers not only the semantic similarity of the synsets but also the overall structure of the alignment and its consistency, following the best practices in ontology alignment.

This work has led to the development of a large multilingual WordNet in more than 20 European languages, which we call Polylingual WordNet \cite{arcan2016expanding},\footnote{\url{http://polylingwn.linguistic-lod.org/}} which is available under an open (CC-BY) license. Finally, we describe how this resource is published, firstly as linked data in the linguistic linked open data cloud,\footnote{\url{http://linguistic-lod.org/llod-cloud}} as well as published in all the formats of the Global WordNet Association Interlingual Index.

\subsection{The Languages covered in Polylingual Wordnet}


The Princeton WordNet is one of the most important resources for natural language processing, but is only available for English. While it has been translated using the \textit{expand} approach to many other languages, this process is a very time consuming and expensive process. Therefore we engage SMT to automatically translate WordNet entries in to 23 European languages, as seen in Table~\ref{tab:eu_lang}. With this amount of languages, Polylingual Wordnet covers eight different language families, i.e. Slavic, Germanic, Uralic, Romance, Hellenic, Celtic, Baltic and Semitic. Furthermore, the entries in the described wordnet are, besides the Latin script, represented in Cyrillic for Bulgarian and Greek alphabet for the Greek language.

\begin{table}
\centering
\setlength{\tabcolsep}{9pt}
\begin{tabular}{lllll}
\toprule
Bulgarian & Estonian  & Hungarian  &  Maltese &  Slovene \\
Croatian & Finnish  & Irish  & Polish  &  Spanish \\
Czech &  French & Italian  &  Portuguese &  Swedish \\
Danish & German  & Latvian  &  Romanian \\
Dutch & Greek & Lithuanian &  Slovak \\
\bottomrule
\end{tabular}
\caption{Languages covered in the Polylingual Wordnet.}
\label{tab:eu_lang}
\end{table}

\subsection{Development of Polylingual WordNet}





 The Princeton WordNet is a large, publicly available lexical semantic database of English nouns, verbs, adjectives and adverbs, grouped into synsets ($\approx$ 117,000), which are aligned in terms of semantic and lexical relations. While it has been translated using the \textit{expand} approach to many other languages, this is an expensive manual process. Therefore it would be beneficial to have a high-quality automatic translation approach that would support NLP techniques, which rely on WordNet in new languages. The translation of wordnets is fundamentally complex because of the need to translate all senses of a word including low frequency senses, which is very challenging for current machine translation approaches. 


Our approach takes the advantage of the increasing amount of parallel corpora in combination with wordnets in languages other than English for sense disambiguation, which helps us to improve automatic translations of English WordNet entries. 
We assume that we have a multilingual parallel corpus consisting of sentences,
$x_i^l$ in a language $l$, grouped into parallel translations:
\[
    \mathcal{X} = \{ (x_i^{l_0}, \ldots, x_i^{l_T}) \}
\]
We also assume that we have a collection of wordnets consisting of a set of
senses, $w_{ij}^l$, grouped into synsets, for each language:
\[
    \mathcal{S} = \{ (\{w_{ij}^{l_0}\}, \ldots, \{w_{ij}^{l_T}\} ) \}
\]
We say that a context $x_i^{l_0}$, in language $l_0$ (in our case this is always
English), is \emph{disambiguated in $n$ languages} for a word $w_{jk}^{l_0}$ if:
\[
    \exists w_{jk_1}^{l_1}, \ldots, w_{jk_n}^{l_n} : w_{jk_1}^{l_1} \in
    x_i^{l_1} \wedge \ldots \wedge w_{jk_n}^{l_n} \in x_i^{l_n}
\]
That is, a context is disambiguated in $n$ languages for a word, if for each of its translations we have a context in the parallel corpus that contains one of the known synset translations. Furthermore, we assume we have an SMT system that can translate any context  in $l_0$ into our target language, $l_T$, and produces an alignment such that we know which word or phrase in the output corresponds to the input. 

Within the set of identified disambiguated contexts, the $m$ top scoring contexts are used, with ties broken at random. Each of these contexts is given to the SMT system and the most frequent translation across these $m$ contexts is used. Furthermore, the SMT system is configured to return the $t$ highest scoring translations, according to its model, and we select the translation as the most frequent translation of the context among this $t$-best list. In our experiments, we combined this with $m$ disambiguations to give $tm$ candidate translations from which the candidate is chosen.

Since only WordNet synsets are linked across different languages, we first align them with its translation equivalents, which is performed with their appearance within several million parallel sentences. In the next step we identify English sentences, which contain an English WordNet entry. Due to the multilingual nature of a parallel corpus, we identify the non-English Wordnet sense on the target side of the parallel corpus. Our approach is based on the assumption that a sentence shares the same semantic information as the WordNet entry sysnset if its translation, with the same mining or synset respectively, appears in the parallel target sentence. This disambiguation approach can be further strengthened, if translations of the targeted WordNet entry appear in several languages in the parallel corpus. Due to this assumption we use 16 different languages in our experiment, which requires 16 different non-English wordnets and parallel corpora. Besides the Princeton Wordnet, we engage wordnets, freely provided by the Open Multilingual Wordnet (OMW) web page,\footnote{\url{http://compling.hss.ntu.edu.sg/omw/}}  i.e.:
\begin{itemize}
    \item Bulgarian~\cite{Simov:Osenova:2010}
    \item Croatian~\cite{Oliver:Sojat:Srebacic:2015}
    \item Danish~\cite{Pedersen2009}
    \item Dutch~\cite{Postma:Miltenburg:Segers:Schoen:Vossen:2016}
    \item Finnish~\cite{Linden:Carlson:2010}
    \item French~\cite{Sagot:Fiser:2008}
    \item Greek~\cite{Stamou:Nenadic:Christodoulakis:2004}
    \item Italian~\cite{Toral:Bracale:Monachini:Soria:2010}
    \item Lithuanian~\cite{Garabik:Pileckyte:2013}
    \item Polish ~\cite{plwordnet2}
    \item Portuguese~\cite{Paiva:Rademaker:2012}
    \item Romanian~\cite{Tufis:Ion:Bozianu:Ceausu:Stefanescu:2008}
    \item Slovak~\cite{horak2004slovak}
    \item Slovene~\cite{Fiser:Novak:Eejavec:2012}
    \item Spanish~\cite{Gonzalez-Agirre:Laparra:Rigau:2012}
    \item Swedish~\cite{Borin:Forsberg:Loenngren:2013}
\end{itemize}

Once we obtain a set of sense disambiguated sentences for each Wordnet entry, we start the translation approach. Our hunch is that correctly identified contextual information around the WordNet entry can guide the SMT system to correctly translate an ambiguous entry.

\subsection{Applications of Polylingual Wordnet} 


Polylingual WordNet was developed as part of the MixedEmotions project,\footnote{\url{https://mixedemotions-project.eu/}} which aims to develop an innovative, multilingual model platform for emotion analysis. As such, we included emotion information from two sources into Polylingual WordNet, namely the emotion analysis from WordNet Affect~\cite{strapparava2004wordnet} and the Emo WordNet data from plWordNet 3.0~\cite{maziarz2016plwordnet}. These have been used for developing emotion classification tools that take the annotations and use them as features in a supervised classification task as part of the SenPy~\cite{sanchez2016senpy} system.

\section{Current state of the Polylingual Wordnet}

This wordnet is currently under active development and we plan to improve the quality of the resource over several iterations. The first release (1.0) was made in November 2016 and a second release has been made with this publication (1.1) in May 2017.









\subsection{Native Statistics} 

The left part of Table~\ref{tab:used_wns} illustrates the size of the wordnets provided by the Open Multilingual Wordnet. 
The right part of the table shows statistics of the Polylingual Wordnet.

\begin{table}
\setlength{\tabcolsep}{20.9pt}
\centering
\begin{tabular}{lccc}
\toprule
& \multicolumn{3}{c}{Princeton Wordnet}\\
\midrule
Language & Synsets & Words & Senses\\
\midrule
English & 117,659 & 206,941 & 147,306 \\
\bottomrule
\end{tabular}
\setlength{\tabcolsep}{3.9pt}
\centering
\begin{tabular}{lccc||lccc}
\toprule
& \multicolumn{3}{c}{Open Multilingual Wordnet} & \multicolumn{3}{c}{Polylingual WordNet} \\
\midrule
Language & Synsets & Words & Senses & Synsets & Words & Senses\\
\midrule
Bulgarian & 4,959 & 6,720 & 8,936 &  117,587 & 235,256 & 140,472 \\
Croatian & 23,120 & 29,008 & 47,900 & 117,634 & 202,343 & 132,630 \\ 
Czech  & / & / & / & 117,563 & 237,998 & 148,141 \\
Danish & 4,476 & 4,468 & 5,859 & 117,516 & 249,204 & 148,384 \\
Dutch & 30,177 & 43,077 & 60,259 & 117,538 & 198,985 & 120,899 \\
Estonian  & / & / & / & 117,607 & 329,687 & 212,263 \\
Finnish & 116,763 & 129,839 & 189,227 & 116,906 & 202,600 & 121,545 \\ 
French & 59,091 & 55,373 & 102,671 & 117,530 & 199,805 & 127,561 \\
German & / & / & / & 117,614 & 225,608 & 144,632 \\
Greek & 18,049 & 18,227 & 24,106 & 117,406 & 217,974 & 127,967 \\
Hungarian & / & / & / & 117617 & 230,442 & 144,086 \\
Irish & / & / & / & 117,571 & 212,183 & 142,230 \\
Italian  & 35,001 & 41,855 & 63,133 & 117,267 & 198,366 & 119,545 \\
Latvian & / & / & / & 117,629 & 203,814 & 136,599 \\
Lithuanian & 9,462 & 11,395 & 16,032 & 117,578 & 203,246 & 135,170 \\
Maltese & / & / & / & 117,647 & 205,406 & 143,248 \\
Polish & 33,826 & 45,387 & 52,378 & 117,071 & 199,322 & 121,743 \\ 
Portuguese & 43,895 & 54,071 & 74,012 & 117,543 & 198,416 & 120,368 \\
Romanian & 56,026 & 49,987 & 84,638 & 117,471 & 199,377 & 121,295 \\ 
Slovak & 18,507 & 29,150 & 44,029 & 117,564 & 203,317 & 131,170 \\
Slovene & 42,583 & 40,233 & 70,947 & 117,594 & 233,854 & 142,799 \\
Spanish & 38,512 & 36,681 & 57,764 & 117,572 & 199,648 & 127,006 \\
Swedish & 6,796 & 5,824 & 6,904 & 117,591 & 206,684 & 130,615 \\
\bottomrule
\end{tabular}
\caption{Statistics on existing wordnets and Polylingual Wordnet.}
\label{tab:used_wns}
\end{table}

\subsection{Special Characteristics of Polylingual Wordnet}


Polylingual WordNet is the largest multilingual resource released under an open license and the only resource that has been developed in a fully automatic manner. As such, the resource plays a number of useful roles in covering applications for languages or applications where the manually constructed wordnets do not have sufficient coverage. Moreover, this WordNet is intended to be a basis that can help in the translation of existing wordnets by providing a basis from which lexicographers can work. 

\subsection{Bridging the Language Barrier}
The construction of Polylingual WordNet is based on phrase-based SMT \cite{koehn2003statistical}, where we wish to find the best translation of a string, given by a log-linear model combining a set of features. The translation that maximizes the score of the log-linear model is obtained by searching all possible translations candidates. The decoder, which is a search procedure, provides the most probable translation based on a statistical translation model learned from the training data.

\subsubsection{Statistical Machine Translation}
For our translation task, we use the statistical translation toolkit Moses \cite{Koehn:2007}, where word alignments, necessary for generating translation models, were built with the GIZA++ toolkit \cite{Och:2003}. The Kenlm toolkit \cite{Heafield-kenlm} was used to build a 5-gram language model.

To ensure a broad lexical and domain coverage of our SMT system we merged the existing parallel corpora for each language pair from the OPUS web page\footnote{\url{http://opus.lingfil.uu.se/index.php}} into one parallel data set, i.e., Europarl \cite{Koehn:2005}, DGT - translation memories generated by the \textit{Directorate-General for Translation} \cite{DBLP:journals/lre/SteinbergerEPCSPG14}, MultiUN corpus \cite{MultiUN}, EMEA, KDE4, OpenOffice~\cite{Tiedemann2009}, OpenSubtitles2012~\cite{Tiedemann:EACL12}, among others.

\subsubsection{Translation Evaluation Metrics}
The automatic translation evaluation is based on the correspondence between the SMT output and reference translation (gold standard). For the automatic evaluation we used the BLEU~\cite{Papineni:2002}, METEOR \cite{denkowski:lavie:meteor-wmt:2014} and chrF~\cite{popovic:2015:WMT} metrics.

\textbf{BLEU} (BiLingual Evaluation Understudy) is calculated for individual
translated segments (n-grams) by comparing them with a data set of reference
translations. 
The calculated scores, between 0 and 100 (perfect translation), are averaged over the whole \textit{evaluation data set} to reach an estimate of the translation's overall quality. Considering the short length of the terms in WordNet, while we report scores based on the unigram overlap (BLEU-1), and as this is in most cases only precision, so in addition we also report other metrics.  

\textbf{METEOR} (Metric for Evaluation of Translation with Explicit ORdering) is based on the harmonic mean of precision and recall, whereby recall is weighted higher than precision. In addition to exact word (or phrase) matching it has additional features, i.e. stemming, paraphrasing and synonymy matching. In contrast to BLEU, the metric produces good correlation with human judgement at the sentence or segment level. 

\textbf{chrF3} is a character n-gram metric, which has shown
very good correlations with human judgements on the WMT2015 shared metric
task~\cite{stanojevic-EtAl:2015:WMT}, especially when translating from English into morphologically rich(er) languages. As there are multiple translations available for each sense in the target wordnet we use all translations as multiple references for BLEU, for the other two metrics we compare only to the most frequent member of the synset.


\subsubsection{Automatic Translation Evaluation}

Here we present the evaluation of the translated English WordNet words into 16 European languages. We evaluate the quality of translations of the WordNet entries against the existing entries in the non-English Wordnets (Table~\ref{tab:smt_eval}).

\begin{table}
\setlength{\tabcolsep}{10pt}
\centering
\begin{tabular}{lcccr}
\toprule
& BLEU-1 & METEOR & chrF & \# Senses \\
\midrule
Bulgarian & 21.0 & 12.8 & 45.3 & 13,264\\
Croatian & 21.7 & 13.4 & 40.3 & 46,729 \\
Danish & 22.3 & 16.3 & 36.9 & 12,481 \\ 
Dutch & 20.7 & 15.0 & 40.5 & 57,970 \\
Finnish & 20.7 & 12.5 & 36.9 & 201,349 \\ 
French & 33.4 & 22.4 & 52.3 & 115,103\\
Greek & 22.8 & 26.0 & 41.0 & 39,546 \\
Italian  & 22.8 & 14.2 & 41.4 & 67,919\\
Lithuanian  & 19.0 & 11.1 & 37.2 & 16,223 \\
Polish & 17.3 & 10.3 & 33.3 & 63,975 \\ 
Portuguese & 32.0 & 19.1 & 48.7 & 84,531 \\
Romanian & 13.2 & 8.8 & 39.1 & 105,342\\ 
Slovak & 23.0 & 11.4 & 36.1 & 33,721\\
Slovene & 18.1 & 10.3 & 38.5 & 103,670 \\
Spanish & 37.0 & 21.3 & 51.2 & 71,765 \\
Swedish & 24.7 & 18.0 & 38.8 & 15,623\\
\bottomrule
\end{tabular}
\caption{Evaluation of WordNet translations into several European languages with context-aware techniques (\# Senses = number of wordnet entries used for evaluation).}
\label{tab:smt_eval}
\end{table}

\subsection{Links to external resources}

Polylingual WordNet contains links to DBpedia (and hence also to Wikipedia), which as with the nature of this resource have been automatically constructed using a dataset alignment approach.

\subsubsection{Automatic linking}

In order to link Polylingual WordNet to other resources, we really on a dataset alignment system called
NAISC (Nearly Automatic Alignment of SChema),\footnote{Pronounced `nashk' and means `links' in Irish Gaelic} which is a system designed to create linking between two resources. As such NAISC takes two lists of entities as input that may have the following:

\begin{description}
    \item[Labels] Each entity may have multiple labels in multiple languages, which are simply strings. In the case of wordnets, these are the words in each synset.
    \item[Descriptions] As labels these are strings grouped by language, and correspond to synset definitions in wordnet.
    \item[Relations] A list of relations to other entities. The list of relation types is fixed, and corresponds well to those in wordnets, e.g., `broader' $\rightarrow$ `hypernym'.
    \item[Type] The type of an entity, in the case of wordnet this is a part-of-speech.
\end{description}

The development of a match is performed by a \emph{matcher}, which defines the process for finding the 
optimal match between two lists of entities. The most straightforward matcher is called the \emph{greedy}
matcher, which simply compares each pair of entities between the two datasets using a similarity 
function to compute the \emph{similarity} between this pair. The similarity calculation is divided into three stages: firstly, the \emph{lens} examines the entity pair and extracts information that can be more easily compared, for example a pair of labels one for each element. This is then fed into a \emph{feature extractor} 
that analyses the facet to produce a numerical value that is assumed to be related to the similarity of this entity pair. Finally, a supervised \emph{similarity classifier} is used to aggregate a number of features
and is trained on existing training data. This then creates a single score between zero (totally dissimilar) and one (identical) for each pair of entities between the two datasets. The \emph{greedy matcher} then 
proceeds by adding matches between the dataset that increase a \emph{topological constrained score} (TCS), which 
is a function that both checks the validity and computes the score of a matching between two datasets. In our 
experiments we use the \emph{bipartite} TCS, which constraints the matching such that no entity in either dataset is linked twice and then produces as a score, the sum of all similarity scores in the dataset. This is summarized in the algorithm in Figure~\ref{algo:naisc}.

\begin{figure}
\begin{algorithmic}
\STATE \textbf{Input:} Two lists of elements (synsets) $D_1$, $D_2$
\STATE \textbf{Output:} A matching between these datasets
\STATE \textbf{Configuration:} A list of \emph{lenses}, list of \emph{extractors}, a similarity functions (\emph{sim}) and a matching \emph{constraint}

\FOR{$d_1 \in D_1$, $d_2 \in D_2$}{}
    \FOR{$\text{lens} \in \text{lenses}$}{}
      \STATE facet $\leftarrow$ lens.apply($d_1$, $d_2$)
      \FOR{$\text{extractor} \in \text{extractors}$}
         \STATE features$_i$ $\leftarrow$ extractor.extract(facet)
      \ENDFOR
    \ENDFOR
    \STATE{sim($d_1$,$d_2$) $\leftarrow$ sim(features)}
\ENDFOR
\STATE score $\leftarrow$ 0
\FOR{$(d_1,d_2) \in D_1 \times D_2$ sorted by sim}{}
    \IF{constraint.can\_add($d_1$,$d_2$,matches)}  
        \STATE matches.add($d_1$,$d_2$) 
        \STATE score $\leftarrow$ score + sim($d_1$,$d_2$) 
    \ENDIF
\ENDFOR
\RETURN matches, score
\end{algorithmic}
\caption{\label{algo:naisc} The NAISC ``greedy constrained'' algorithm for matching datasets.}
\end{figure}

In order to apply NAISC to the task of matching WordNet to Wikipedia, we need to define a set of lenses and 
feature extractors to calculate similarity between these entities. The following lenses were used in our experiments

\begin{description}
    \item[Most similar labels] We extract every word from the target WordNet synset and for Wikipedia, the 
    article title and all other titles that redirect to this article. We choose the pair that has the lowest
    Levenshtein distance between each other.
    \item[Concatenated labels] We extract labels as above for WordNet and Wikipedia, but instead concatenate all the words extracted
    \item[Description] We compare the definition of the concept for WordNet with the first sentence of the Wikipedia article
    \item[Superterms] For each WordNet synset we extract all the words in every concept that is a hypernym (either directly or transitively). For Wikipedia we used the assigned categories in the DBpedia Ontology and YAGO Ontology~\cite{suchanek2007yago}. We use the two labels extracted in this way that are closest in Levenshtein distance.
\end{description}

All of the lenses that we extract are textual and as such we apply the following textual features, which were combined into a single similarity score using Weka's SMO classifier~\cite{Platt1998}.

\begin{description}
\item[Jaccard, Dice, Containment] We consider the two strings both as a set of words and a set of characters and compute the following functions $J(A,B) = |A \cap B| / |A \cup B|$, $D(A,B) = 2|A \cap B| / (|A| + |B|)$,
  $C(A,B) = |A \cap B| / \min(|A|,|B|)$.
\item[Smoothed Jaccard] Smoothed Jaccard is calculated only on the word level for the concatenated labels facet. It is calculated as follows $J_\sigma(A,B) = \sigma(|A \cap B|) / (\sigma(|A|) + \sigma(|B|) - \sigma(|A \cap B|))$ where $\sigma(x) = 1.0 - exp(-\alpha x)$. This is a variant of Jaccard that can be adjusted to distinguish matches on shorter texts; it tends to Jaccard at $\alpha \rightarrow 0$.
\item[Length Ratio] The ratio of the number of tokens in each sentence. For symmetry this ratio is defined as 
 $\rho(x,y) = \frac{\min(x,y)}{\max(x,y)}$.
\item[Average Word Length Ratio] The average length of each word in the text are also compared as above.
\item[Negation] One if both texts or neither text contain a negation word (`not', `never', etc.), zero otherwise.
\item[Number] One if all numbers (e.g, `6') in each text are found in the other, zero otherwise.
\item[GloVe Similarity] For each word in each text we extract the GloVe vectors~\cite{pennington2014glove} and calculate the cosine similarity between these words, to give a value $\sigma_{ij}$ between the $i$\textsuperscript{th} and the $j$\textsuperscript{th} word. We calculate a score as:
\begin{equation*}
    \frac{1}{n}\sum_{i = 1, \ldots n} \max_{j = 1, \ldots n} \sigma_{ij}
\end{equation*}
where $n$ is the length of the WordNet text and $m$ is the length of the Wikipedia text.
\item[LSTM] We calculate a similarity using the LSTM approach described by~\cite{tai2015improved}.
\end{description}

As each of these features can be turned on or off individually we consider the effects of each for the quality
of the alignment between Princeton WordNet and Wikipedia. For this we used the 200NS dataset from \cite{fernando2012mapping}, and computed the precision, recall and F-measure of the mapping for various settings, using 10-fold cross validation, as follows:

\begin{enumerate}
    \item Only Jaccard score of concatenated label
    \item Only smoothed Jaccard ($\alpha=1$) of concatenated label
    \item As above with \emph{basic} features (Jaccard, Dice, Containment, Length Ratio, Average Word Length Ratio, Negation and Number) for the most similar label
    \item As above with \emph{basic} features for label and description
    \item As above with GloVe Similarity
    \item As above with LSTM score
    \item As 5, but using superterm labels
    \item As 7, but using superterm labels
\end{enumerate}

\begin{table}
\begin{center}
\begin{tabular}{lccc}
\toprule
Setting & Precision & Recall & F-Measure \\
\midrule
1       & 38.5\%    & 46.2\% & 42.0\%    \\
2       & 40.1\%    & 54.9\% & 46.3\%    \\
3       & 39.8\%    & 54.3\% & 46.0\%    \\
4       & 43.7\%    & 60.1\% & 50.6\%    \\
5       & 45.6\%    & 63.0\% & 52.9\%    \\ 
6       & 45.8\%    & 63.0\% & 52.9\%    \\
7       & 45.8\%    & 63.0\% & 53.0\%    \\
8       & 46.2\%    & 63.6\% & 53.5\%    \\
\bottomrule
\end{tabular}
\end{center}
\caption{\label{tab:naisc-results}Results for the NAISC system on aligning WordNet and Wikipedia.}
\end{table}

The results show that the combination of different factors improves the matching quality and we see large gains in the score by combining these features. As such this was used as a matching algorithm to create the final mapping between Princeton WordNet and DBpedia.

\section{Related Work}
\label{sec:related}

The Princeton WordNet inspired many researchers to create similarly structured wordnets for other languages. The EuroWordNet project \cite{Vossen:1998:EMD:314515} linked  wordnets in different languages through a so-called Inter-Lingual-Index (ILI) into a single multilingual lexical resource. Via this index, the languages are aligned between each other, which allows to go from a concept in one language to a concept with a similar meaning in any of the other languages. Further multilingual extensions were generated by the BalkaNet project \cite{tufis2004balkanet}, focusing on the Balkan languages and MultiWordNet \cite{MultiWordNet_2002}, aligning Italian concepts to English equivalents.

Due to the large interest in the multilingual extensions of the Princeton WordNet, several initiatives started with the aim to unifying and making these wordnets easily accessible. The KYOTO  project \cite{Fellbaum:2012:CMW:2386778.2386832} focused on the development of a language-independent module to which all existing wordnets can be connected, which would allow a better cross-lingual machine processing of lexical information. Recently this has been realized by a new Global WordNet Grid \cite{vossen2016toward} that takes advantage of the Collaborative Inter-Lingual Index (CILI) \cite{bond2016cili}. Since most of the current non-English wordnets use the Princeton WordNet as a pivot resource, concepts, which are not in this English lexical resource cannot not be realized or aligned to it. Therefore the authors support the idea of a central platform of concepts, where new concepts may be added even if they are not represented (yet) in the Princeton WordNet or even lexicalized in English (e.g., many languages have distinct gendered role words, such as `male teacher' and `female teacher', but these meanings are not distinguished in English).
 
Previous studies of generating non-English wordnets combined Wiktionary knowledge with existing wordnets to extend them or to create new ones~\cite{deMeloWeikum2009}.

\cite{BondPaik2012} describe in their work the creation of the Open Multilingual Wordnet and its extension with other resources \cite{Bond:Foster:2013}. The resource is made by combining different wordnets together, knowledge from Wiktionary and the Unicode Common Locale Data Repository. Overall they obtained over 2 million senses for over 100 thousand concepts, linking over 1.4 million words in hundreds of languages. Since using existing lexical resources guarantees a high precision, it may also provide a low recall due to the limitedness of lexical resources in different languages and domains.
A different approach to expand English WordNet synsets with lexicalizations in other languages was proposed in \cite{deMelo2012}. The authors do not directly match concepts in the two different language resources, but demonstrate an approach that learns how to determine the best translation for English synsets by taking bilingual dictionaries, structural information of the English WordNet and corpus frequency information into account. With the growing amount of parallel data, \cite{Kazakov:2009:UCM:1859657.1859659} show an approach to acquire a set of synsets from parallel corpora. The synsets are obtained by comparing aligned words in parallel corpora in several languages. Similarly, the sloWNet for Slovene \cite{fivser2007leveraging} and Wolf for French \cite{Sagot:Fiser:2008}
are constructed using a multilingual corpus and word alignment techniques in
combination with other existing lexical resources. 
Since all these approaches use word alignment information, they are not able to generate any translation equivalents for multi-word expressions (MWE). In contrast, our approach use an SMT system trained on a large amount of parallel sentences, which allows us to align possible MWEs, such as \textit{commercial loan} or \textit{take a breath}, between source and target language. Furthermore, we engage the idea of identifying relevant contextual information to support an SMT system translating short expressions, which showed better performance compared to approaches without a context. \cite{arcan2015:ACL} built small domain-specific translation models for ontology translation from relevant sentence pairs that were identified in a parallel corpus based on the ontology labels to be translated. With this approach they improve the translation quality over the usage of large generic translation models. Since the generation of translation models can be computational expensive, \cite{arcan2016iswc} use large generic translation models to translate ontology labels, which were placed into a disambiguated context. With this approach the authors demonstrate translation quality improvement over commercial systems, like \textit{Microsoft Translator}. Different from this approach, which uses the hierarchical structure of the ontology for disambiguation, we engage a large number of different languages to identify the relevant context.

\cite{DBLP:conf/cicling/OliverC12} present a method for WordNet construction and enlargement with the help of sense tagged parallel corpora. Since parallel sense tagged data are not always available, they use \textit{Google Translate} to translate a manually sense tagged corpus. In addition they apply automatic sense tagging of a manually translated parallel corpus, whereby they report worse performance compared to the previous approach. We try to overcome this issue by engaging up to ten languages to improve the performance of the automatic sense tagging. Similarly, BabelNet \cite{Navigli:2012:BAC:2397213.2397579} aligns the lexicographic knowledge from WordNet to the encyclopaedic knowledge of Wikipedia. This is done by assigning WordNet synsets to Wikipedia entries, and making these relations multilingual through the interlingual links. For languages, which do not have the corresponding Wikipedia entry, the authors use \textit{Google Translate} to translate English sentences containing the synset in the sense annotated corpus. After that, the most frequent translation is included as a variant for the synset for the given language.

The use of parallel corpora has been previously exploited for word sense disambiguation, for example to construct sense-tagged corpora in another language~\cite{ng2003exploiting} or by using translations as a method to discriminate senses~\cite{ide2002sense}. It has been shown that the combination of these techniques can improve supervised word sense disambiguation~\cite{chan2007nus}. A similar approach to the one proposed in this paper is that of \cite{tufics2004fine}, where they show that using the interlingual index of WordNet with the help of parallel text can improve word sense disambiguation of a monolingual approach and we generalize this result to generate wordnets for new languages. 

\section{Discussion and Future Plans}


Polylingual WordNet is a wordnet that has been constructed fully automatically based on the English Princeton WordNet and as such represents a significantly different resource to the others described in this volume, yet a resource that will still be helpful in a wide number of applications. This resource was created by means of a novel machine translation approach, which uses disambiguated contexts to find the correct translation of a given sense, and has been shown~\cite{arcan2016expanding} that this is significantly better than direct translation. Furthermore, we have also used automatic methods to provide links from this resource to other resources by means of semantic and structural similarity, which gives a high quality linking to encyclopaedic resources, in particular DBpedia/Wikipedia. Thus, while our results show that this resource is of noticeably lower quality than manually constructed resources, there are still many applications where the wide coverage of this resource would be preferred to smaller, high-quality wordnets. We intend to continue to refine our processes, in order to close the gap between this automatically constructed wordnet and manually constructed wordnets in terms of quality. Furthermore, we are working on expanding the coverage of this resource beyond European languages and in particular into under-resourced languages such as Dravidian and Gaelic languages.

\bibliographystyle{unsrt}  
\bibliography {references}

\end{document}